\documentclass[11pt]{article}
\usepackage[T1]{fontenc}
\usepackage[english]{babel}
\usepackage{CJKutf8} 
\usepackage{units}
\usepackage{amsthm}
\usepackage[flushleft]{threeparttable}
\usepackage{rotating}
\usepackage{arydshln}
\usepackage{amsmath}
\usepackage[space]{grffile}
\usepackage{graphicx}
\usepackage{setspace}
\usepackage{esint}
\usepackage{textcomp}
\usepackage{float}
\usepackage{apacite}
\usepackage{multirow}
\usepackage{rotfloat}
\usepackage{xr-hyper}
\usepackage{booktabs}
\usepackage{calc}
\usepackage{CJK}
\usepackage{psfrag,color} 
\usepackage{nicefrac} 
\usepackage{amsfonts} 
\usepackage{subfig} 
\usepackage{ae} 
\usepackage{aecompl} 
\usepackage{pifont} 
\usepackage{natbib} 
\usepackage{hyperref} 
\usepackage{soul}
\usepackage{lscape}
\usepackage{indentfirst}
\usepackage{rotating}
\usepackage[a4paper]{geometry}
\usepackage[flushleft]{threeparttable}
\makeatletter
\theoremstyle{plain}
  
\theoremstyle{plain}

  \theoremstyle{plain}
  
    \theoremstyle{plain}
  
\usepackage{listings}
\usepackage{listings}
\usepackage{xcolor}

\definecolor{codegreen}{rgb}{0,0.6,0}
\definecolor{codegray}{rgb}{0.5,0.5,0.5}
\definecolor{codepurple}{rgb}{0.58,0,0.82}
\definecolor{backcolour}{rgb}{0.95,0.95,0.92}

\lstdefinestyle{mystyle}{
    backgroundcolor=\color{backcolour},   
    commentstyle=\color{codegreen},
    keywordstyle=\color{magenta},
    numberstyle=\tiny\color{codegray},
    stringstyle=\color{codepurple},
    basicstyle=\ttfamily\footnotesize,
    breakatwhitespace=false,         
    breaklines=true,                 
    captionpos=b,                    
    keepspaces=true,                 
    numbers=left,                    
    numbersep=5pt,                  
    showspaces=false,                
    showstringspaces=false,
    showtabs=false,                  
    tabsize=2
}

\title{A Tutorial on the Pretrain-Finetune Paradigm for Natural Language Processing\footnote{Our replication package for this tutorial, with code, data and two additional practical exercises, is available at https://doi.org/10.7910/DVN/QTT84C.}}

\author{Yu Wang and Wen Qu\\Institute for Advanced Study in Social Sciences \\ Fudan University}

\date{}

\begin{document}

\maketitle

\begin{abstract}

\noindent Given that natural language serves as the primary conduit for expressing thoughts and emotions, text analysis has become a key technique in psychological research. It enables the extraction of valuable insights from natural language, facilitating endeavors like personality traits assessment, mental health monitoring, and sentiment analysis in interpersonal communications. In text analysis, existing studies often resort to either human coding, which is time-consuming, using pre-built dictionaries, which often fails to cover all possible scenarios, or training models from scratch, which requires large amounts of labeled data. In this tutorial, we introduce the pretrain-finetune paradigm. The pretrain-finetune paradigm represents a transformative approach in text analysis and natural language processing. This paradigm distinguishes itself through the use of large pretrained language models, demonstrating remarkable efficiency in finetuning tasks, even with limited training data. This efficiency is especially beneficial for research in social sciences, where the number of annotated samples is often quite limited. Our tutorial offers a comprehensive introduction to the pretrain-finetune paradigm. We first delve into the fundamental concepts of pretraining and finetuning, followed by practical exercises using real-world applications. We demonstrate the application of this paradigm across various tasks, including multi-class classification and regression, and illustrate its superior performance compared to traditional methods and GPT-based approaches. Emphasizing its efficacy and user-friendliness, the tutorial aims to encourage broader adoption of this paradigm. To this end, we have provided open access to all our code and datasets. The tutorial is highly beneficial across various psychology disciplines, providing a comprehensive guide to employing text analysis in diverse research settings.


\end{abstract}



\doublespacing
\setlength{\parindent}{2em}
\section{Introduction}

\noindent The language that we use for expressing ourselves contains rich information on psychological constructs~\citep{text_package,pennebaker2003psychological}. Compared with close-ended numerical rating scales, it has been argued that text data has a higher ecological and face validity~\citep{beyond_rating_scales,semantic_measures}. The technique of translating psychological constructs contained in our language into scientific
measurable units (i.e., words into numbers) is often referred to as text analysis or psychological language processing~\citep{natural_language_analysis}. This technique has been applied to various topics, such as personality assessment~\citep{kwantes2016assessing}, affective beliefs toward physical activity~\citep{affective_beliefs}, abstract screening~\citep{screening,screening_2}, consumers' purchasing intentions~\citep{purchasing_intents}, psychological underpinnings of large organizations' environmental initiatives~\citep{pm}, emotion detection~\citep{emotion_detection} and depression monitoring~\citep{monitor_depression}.

A shared theme woven through these studies is that they invariably attempt to analyze unstructured textual data to understand psychological states and traits, such as in expressions like ``Why don't you ever text me!''~\citep{emotion_detection,natural_language_analysis}. Such data is ubiquitous: whether it is on social media, online 
 forums and blogs, clinical interviews, public speeches, open-ended surveys, diaries, or online reviews of products and services. In psychological research, analyzing these texts is essential for evaluating constructs such as emotional valence, mood states, mindset types, and broader psychological phenomena like resilience, stress levels, and signs of psychological disorders.

To understand these contents, existing studies often resort to either human coding~\citep{affective_beliefs,purchasing_intents}, which is time-consuming and not scalable, using pre-built dictionaries~\citep{liwc,fine_grained_analysis}, which frequently fails to cover all possible scenarios, or training models from scratch using words as tokens or as word embeddings~\citep{screening,screening_2,pm}, which requires large amounts of labeled data and consequently a lot of time to curate. These existing paradigms of text analysis are now being complemented by a rising new paradigm: the pretrain-finetune paradigm.\footnote{For a historical review of psychological language analysis, readers can refer to~\cite{qta} and~\cite{natural_language_analysis}.}

\subsection*{Why the Pretrain-Finetune Paradigm}
\noindent This new paradigm, characterized by the application of large pretrained language models, most notably BERT~\citep{bert} and RoBERTa~\citep{roberta}, and high efficacy on finetuned tasks even in the face of relatively few training samples, is now gaining popularity in social sciences~\citep{monitor_depression,finetune_pa,pretrained_topic_classification}. The reason behind its rising popularity in social sciences is understandable: (1) finetuning large language models is relatively easy to do and virtually all social scientists, regardless of our methodological training, can do it, (2) finetuning large language models requires fewer labeled samples than training models from scratch, and this fits particularly well with social science research where high quality labeled data is scarce, and (3) finetuning large language models is particularly effective, generating state-of-the-art performance in almost all studies that it touches.\footnote{For a detailed introduction of the advantages of language models for psychological research, interested readers could refer to~\cite{beyond_rating_scales}.} As an example, while earlier works have recommended a minimum of 3,000 samples for NLP tasks using the bag-of-words approach~\citep{pm} and various manual feature engineering, recent research applying the pretrain-finetune paradigm has shown that finetuned large models with just a few hundred labeled samples could yield competitive performance with no feature engineering at all~\citep{pretrained_topic_classification}.

\begin{figure}[h]
\centering
\includegraphics[width=410px]{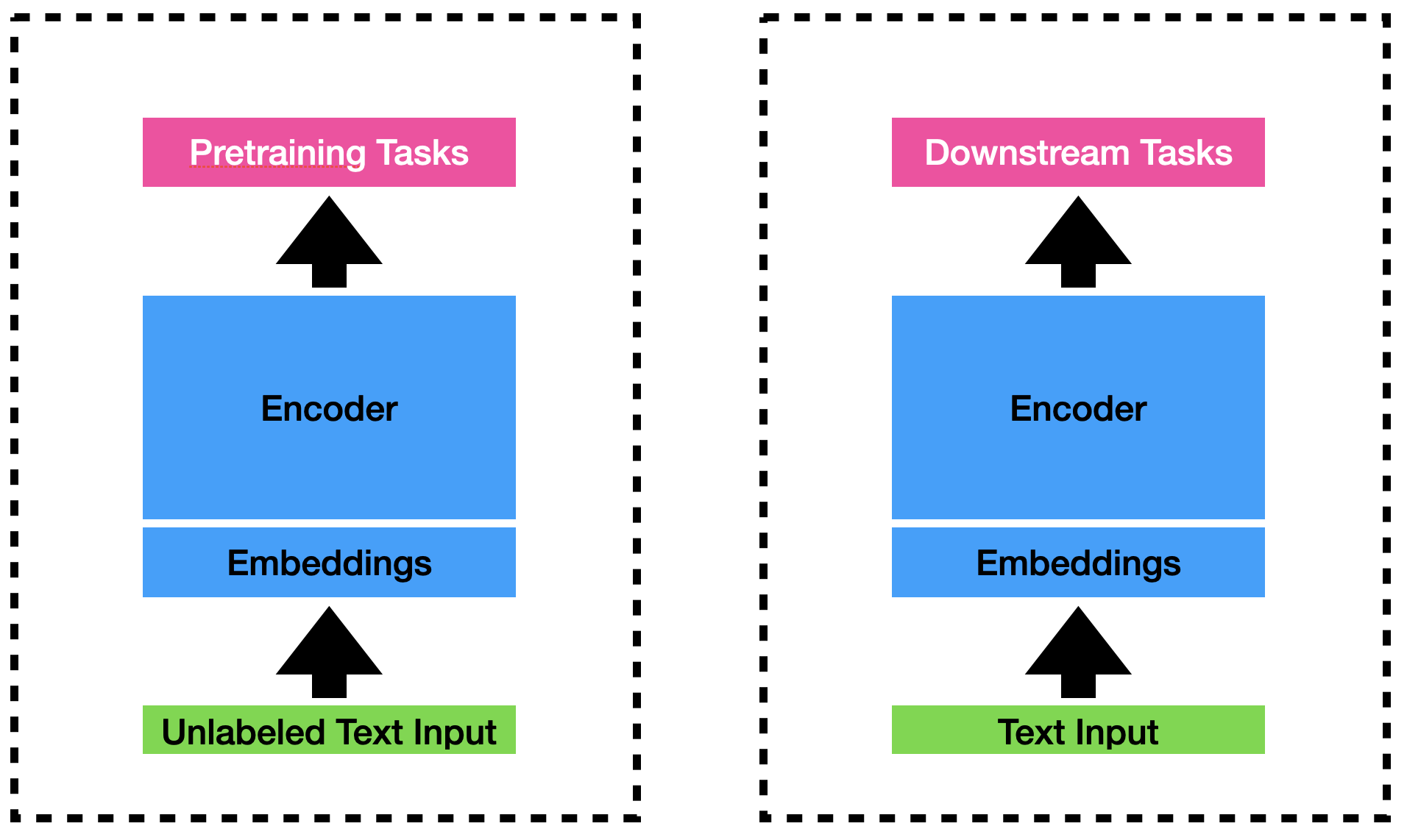}
\caption{High-level illustration of the pretraining (left) and finetuning (right) workflows. Pretraining happens once and is mostly done by large corporations such as Google, Meta, Apple and Microsoft. Finetuning happens whenever a researcher needs to use a pretrained model on a specific task, such as personality classification.}
\label{diagram}
\end{figure}

In the current tutorial, we aim to provide (1) an overview of the pretrain-finetune paradigm for psychologists and (2) illustrative applications of the pretrain-finetune paradigm to research questions in psychology (Figure~\ref{diagram}). We first provide an introduction to the key concepts in pretraining and finetuning, such as tokenization and encoding. We then use four applications, two in the main manuscript and two more in the replication package, of the pretrain-finetune paradigm in social sciences to illustrate how to use this paradigm to advance quantitative research in our discipline. Note that our tutorial covers the pretraining stage mostly for the purpose of completeness as it explains the origins of these pretrained models. Nonetheless, given that we will be using publicly available pretrained models, and that for these models the pretraining stage is already done, our tutorial will focus primarily on the finetuning part: how psychologists could finetune these pretrained models to advance their research.

\section{Pretraining}

\noindent Pretraining is the process of training a model with unlabeled raw data with no particular downstream tasks like sentiment analysis or behavior prediction. Some of the most widely used pretrained models include BERT and RoBERTa. These large language models usually contain hundreds of millions of parameters. Pretraining these models from scratch requires access to large amounts of raw data and specialized hardware like graphics processing units (GPUs). The pretraining process consists of the following steps: (1) tokenization, (2) encoding, and (3) pretraining tasks, such as masked language modeling, next sentence prediction~\citep{bert} and sentence-order prediction~\citep{albert}.

\subsection{Tokenization}
\noindent Unlike earlier methods such as bag of words~\citep{pm} or word embeddings~\citep{word2vec}, large langauge models mostly use subwords as a token, such as wordpieces in BERT~\citep{bert} and Byte-Pair Encoding (BPE) in RoBERTa~\citep{roberta} and GPT-3~\citep{gpt3}. To illustrate how words are broken into subwords, let's use a text snippet from the Journal's description and tokenize it using the BPE tokenizer used in GPT-3.5 and GPT-4.\footnote{The snippet is taken from https://journals.sagepub.com/description/AMP. The tokenizer is available at https://platform.openai.com/tokenizer and was accessed on January 6th 2024.}.


\begin{verbatim}
Psychology is an academic discipline of immense scope, crossing the boundaries
between the natural and social sciences.    
\end{verbatim}



The word ``Psychology'' is split into ``psych'' and ``ology''. All other words, including punctuation marks, remain individual units without further splitting. These units are then turned into tokens represented with non-negative integers.



\begin{verbatim}
[69803, 2508, 374, 459, 14584, 26434, 315, 38883, 7036, 11, 27736, 279,
23546, 1990, 279, 5933, 323, 3674, 36788, 13]
\end{verbatim}


\subsection{Encoding}
\noindent Once we have tokenized the input text into tokens, we then retrieve the corresponding embeddings for these tokens, where the token id, a non-negative integer, would serve the retrieval key. These token embeddings are vector representations. For the BERT-large model, each such vector contains 1024 float numbers. These sets of vectors are then fed into the encoder layers.\footnote{BERT models use encoders and are the focus of this tutorial. Note that GPT models use decoders.}

\begin{figure}[h]
\centering
\includegraphics[width=405px]{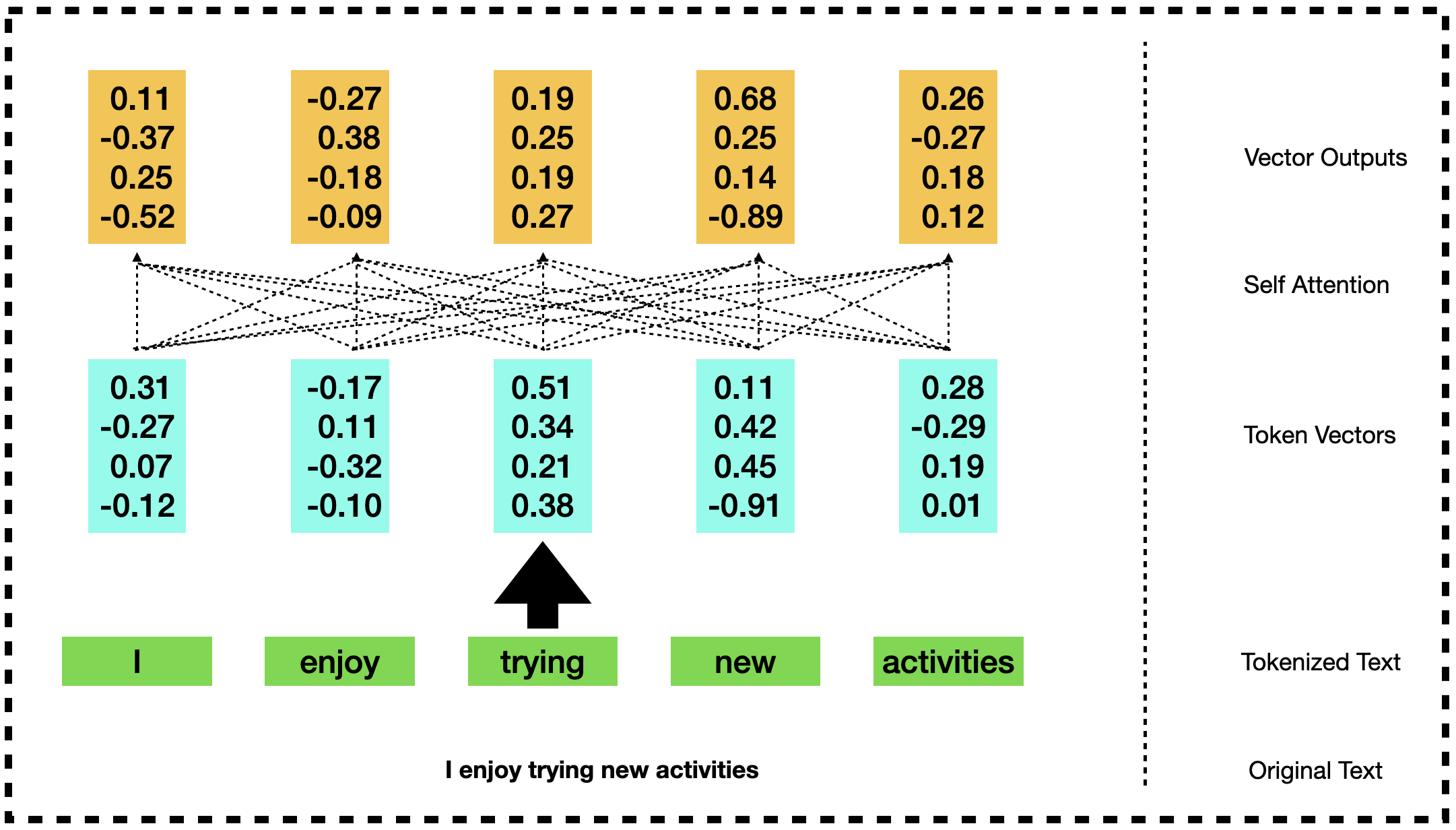}
\caption{Illustration of the BERT model's self-attention mechanism.}
\label{finetune}
\end{figure}

The key component of the encoders is self-attention~\citep{attention}. Intuitively, it computes how much attention each token pays to the other tokens and generates the new representation for a token by calculating the weighted average of all the tokens where the weight is based on self-attention.\footnote{For a similar illustration, readers can refer to~\cite{text_package}. For a more in-depth illustration, please refer to the original paper~\citep{attention} and the tutorial on transformers at http://jalammar.github.io/illustrated-transformer/.} Such encoders are then put on top of each other in a sequential manner. For example, in BERT-large models there are 24 encoder layers. In BERT-base models there are 12 encoder layers.

\subsection{Pretraining tasks}
\noindent Now that we have a good understanding of tokenization and encoding, let's proceed to discuss how these parameters in these layers are trained using pretraining tasks. The goal is to train these large language models from scratch and embue them with world knowledge that exists in the training corpus~\citep{aug}. Commonly used pretraining tasks include masked language modeling, next sentence prediction~\citep{bert} and sentence order prediction~\citep{albert}.\footnote{Later research has shown that the next sentence prediction does not contribute much to the pretraining process~\citep{roberta}.}

\begin{figure}[h]
\centering
\includegraphics[width=395px]{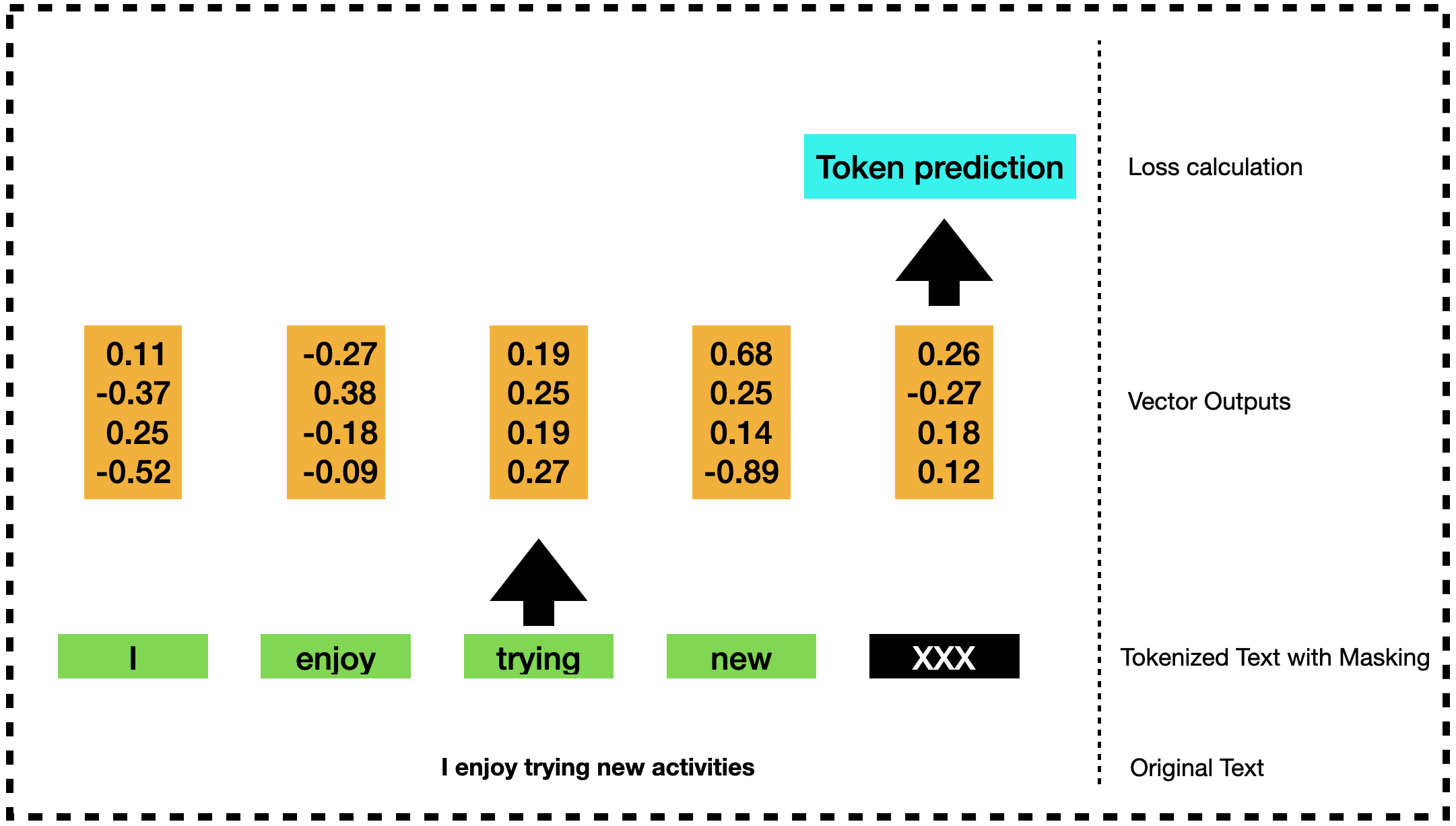}
\caption{Illustration of the BERT model's masked language modeling.}
\label{mlm}
\end{figure}

Let's see an example of how masked language modeling works. Suppose we have the following input text: ``I enjoy trying new activities''. During training, some of the tokens will be randomly masked and the language model is then tasked with predicting these masked tokens. In Figure~\ref{mlm}, the token ``activities'' is masked. The language model then is tasked with predicting this masked token. All tokens in the vocabulary are eligible candidates. Suppose the model predicts ``activities'' with a probability of 0.2. If we use cross entropy loss, then the loss term for this prediction is -log(0.2), which is its negative log-likelihood.\footnote{For details, please see https://github.com/google-research/bert/blob/master/run\_pretraining.py\#L303.} Note that the higher probability the model assigns to the corrected token, the lower the loss. The language model is trained on this task over the entire corpus for a few times until the loss in prediction stops decreasing.

\section{Finetuning}
\noindent Finetuning is the process of training an already-pretrained model on a downstream task. While pretraining is compute-intensive and usually takes place in large corporations and research labs, finetuning requires much less compute and can be carried out by individual researchers. Finetuning is characterized by (1) adding a small set of new parameters to accommodate the new tasks, and (2) a relatively small learning rate given the fact that the vast majority of the parameters are already reasonably trained during the pretraining stage.\footnote{In terms of training procedures, finetuning is the same as training a supervised model from scratch. Readers could refer to a recent tutorial on how to train supervised models by~\cite{best_practices_supervised_ml}.} Downstream tasks at this stage can be broadly grouped into classification and regression. Examples of classification tasks include sentiment analysis and topic classification; examples of regression tasks include predicting the severity of psychological disorder symptoms. This is the step where clinical psychologists apply a pretrained model to a particular task, such as mood disorder detection and personality trait assessment. Similarly, developmental psychologists might finetune a pretrained model to analyze language development stages in children by adjusting the model to recognize and categorize speech patterns from recordings of child interactions. Educational psychologists might use this method to assess learning outcomes from student feedback. In general, for each specific task, researchers need to specify the model's input, targeted output (such as the severity scale or diagnostic category), and whether this is a classification task or a regression task.


\begin{figure}[h]
\centering
\includegraphics[width=210px]{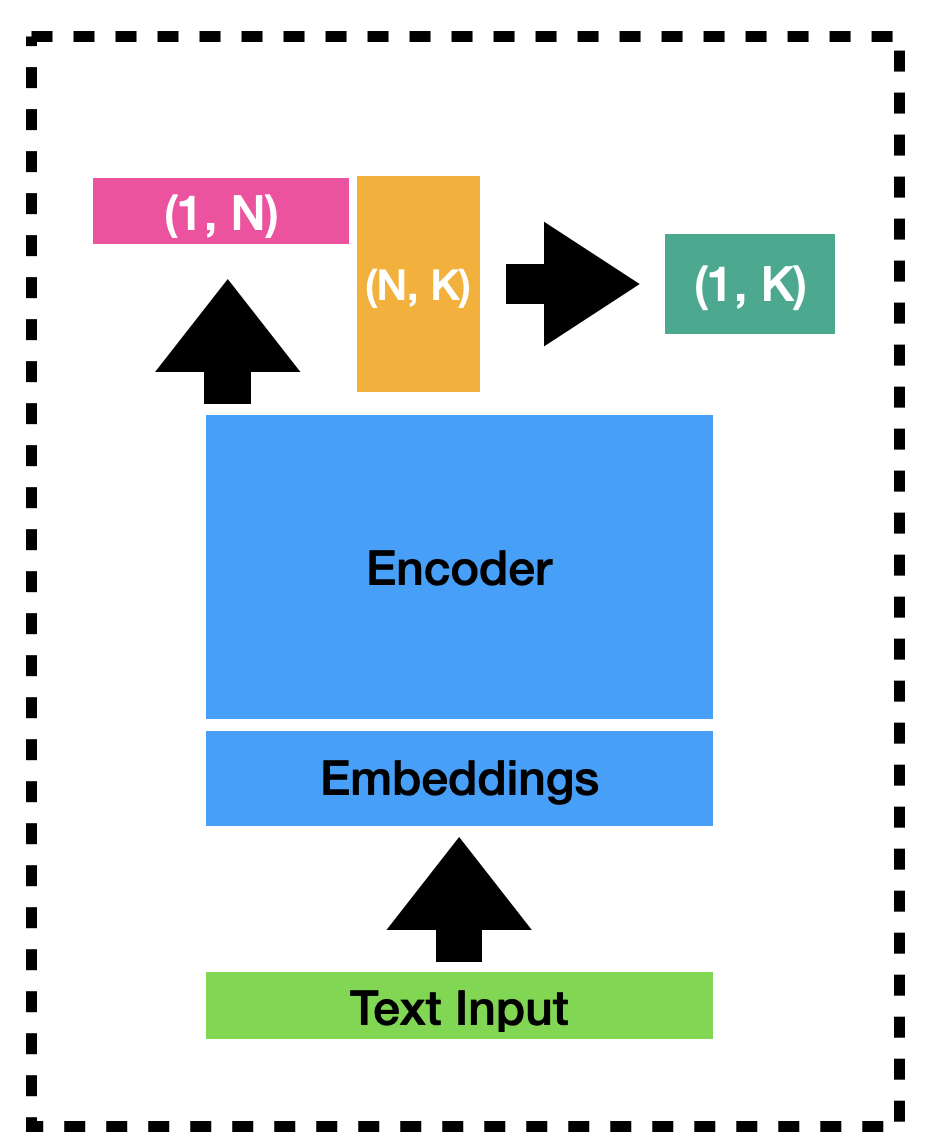}
\caption{Illustration of the finetuning process. Same as with pretraining, the model takes in a text input and generates an encoding for each token. What differentiates finetuning from pretraining is that it takes a particular token representation to make task-specific predictions.}
\label{finetune}
\end{figure}

In Figure~\ref{finetune}, we illustrate how a k-class classification works. From the encoder layers, we retrieve a 1 by N vector for each input sample, where N is 768 in the case of BERT-base and 1024 in the case of BERT-large. Then we project this vector to 1 by K using an N by K matrix. This layer is often referred to as the fully connected layer. What follows next depends on whether the task is classification or regression. In the case of classification, where K is equal to or greater than 2, we apply a softmax function such that each of the K elements represents the probability of that corresponding class being the predicted class:\footnote{For implementation details, refer to https://pytorch.org/docs/stable/generated/torch.nn.Softmax.html.}

\[softmax(x_i) =\frac{exp(x_i)}{\sum_j(exp(x_j))} \] 

\noindent In the case of regression, where K is 1, we can apply the loss function directly to the 1 by K vector (now 1 by 1). Mean squared error is a commonly used loss function for regression tasks.\footnote{For implementation details, refer to https://pytorch.org/docs/stable/generated/torch.nn.MSELoss.html.}

\section{Hyperparameter Tuning}
\noindent When it comes to finetuning language models to downstream tasks, one critical step is selecting appropriate hyperparameters. This is often referred to hyperparameter tuning~\citep{hyperparameters}. Some of the most critical hyperparameters in language models are \textbf{learning rate}, \textbf{batch size}, \textbf{number of training epochs}, and whether or not to use \textbf{mixed precision training}. Learning rate is arguably the single most important hyperparameter when it comes to finetuning~\citep{deep_learning}. When the learning rate is too high, training loss could inadvertently increase rather than decrease. This would be immediately noticeable from the training logs. When the learning rate is too small, learning is slow and ineffective: while the training error will decrease, the pace at which it decreases is too low. In the original BERT paper, the authors selected the best learning rate (among 5e-5, 4e-5, 3e-5, and 2e-5) on the dev set~\citep{bert}. The batch size determines how much data we pack into each training step~\citep{deep_learning}. For example, a training size of 32 would mean that we bundle 32 training samples together as a single input. Given a fixed number of training samples, a larger batch size would mean a smaller number of parameter updates.\footnote{For a discussion on the relationship between the batch size and the learning rate, interested readers can refer to~\cite{batch_size}~\cite{76minutes}, and~\cite{linear_scaling}.} The number of training epochs determines how many times the model goes over the entire dataset. In the original BERT paper, the authors uniformly set the number of training epochs to 3 for all GLUE tasks~\citep{bert}. Mixed precision training~\citep{mixed_precision} stores weights, activation and gradients in half-precision format rather than the default single precision format. It can substantially speed up the training process as it simplifies the calculation. This can become particularly helpful when the dataset becomes large and researchers find that training with full precision is too slow.


\section{Practical Exercises}
\noindent In this section, we provide two practical exercises to illustrate how researchers can transform text into psychological constructs or a continuous variable by finetuning a large language model.\footnote{For more practical exercises, interested readers could refer to the replication package, where we have included two additional examples for multi-class classification and regression.} Our exercises are written in Python in the format of Jupyter notebooks so that readers can follow along in an interactive manner.\footnote{https://jupyter.org.} We will provide the corresponding R implementation if there is sufficient interest. For easy reproducibility and access to computation resources, all our computation is done on Google Colab.\footnote{https://colab.research.google.com.}


\subsection{Multi-class Classification: Comparing Language Models Fine-tuning with Naive Bayes, Max Entropy, Support Vector Machine and Zero-shot Prompting}
\noindent In the following case study, we transform a given text into one of 15 psychological constructs. This represents an example of a typical topic classification task~\citep{pretrained_topic_classification}. Our goal is two-fold: to show how this can be done with relative ease and to show finetuning a large language model is indeed effective and thus worth considering. The dataset we use for this exercise comes from~\cite{pm} and it originated from a research project on environmental sustainability in organizations. In~\cite{pm}, machine learning algorithms, such as naïve bayes, max entropy, and support vector machine (SVM), are used. Note that in~\cite{pm} the authors trained all the models from scratch. In this exercise, we adopt the same dataset and the same evaluation metrics to illustrate how we can use finetuning to do the same thing, i.e., transform texts into psychological constructs, with superior results.

\begin{figure}[h]
\centering
\includegraphics[width=405px]{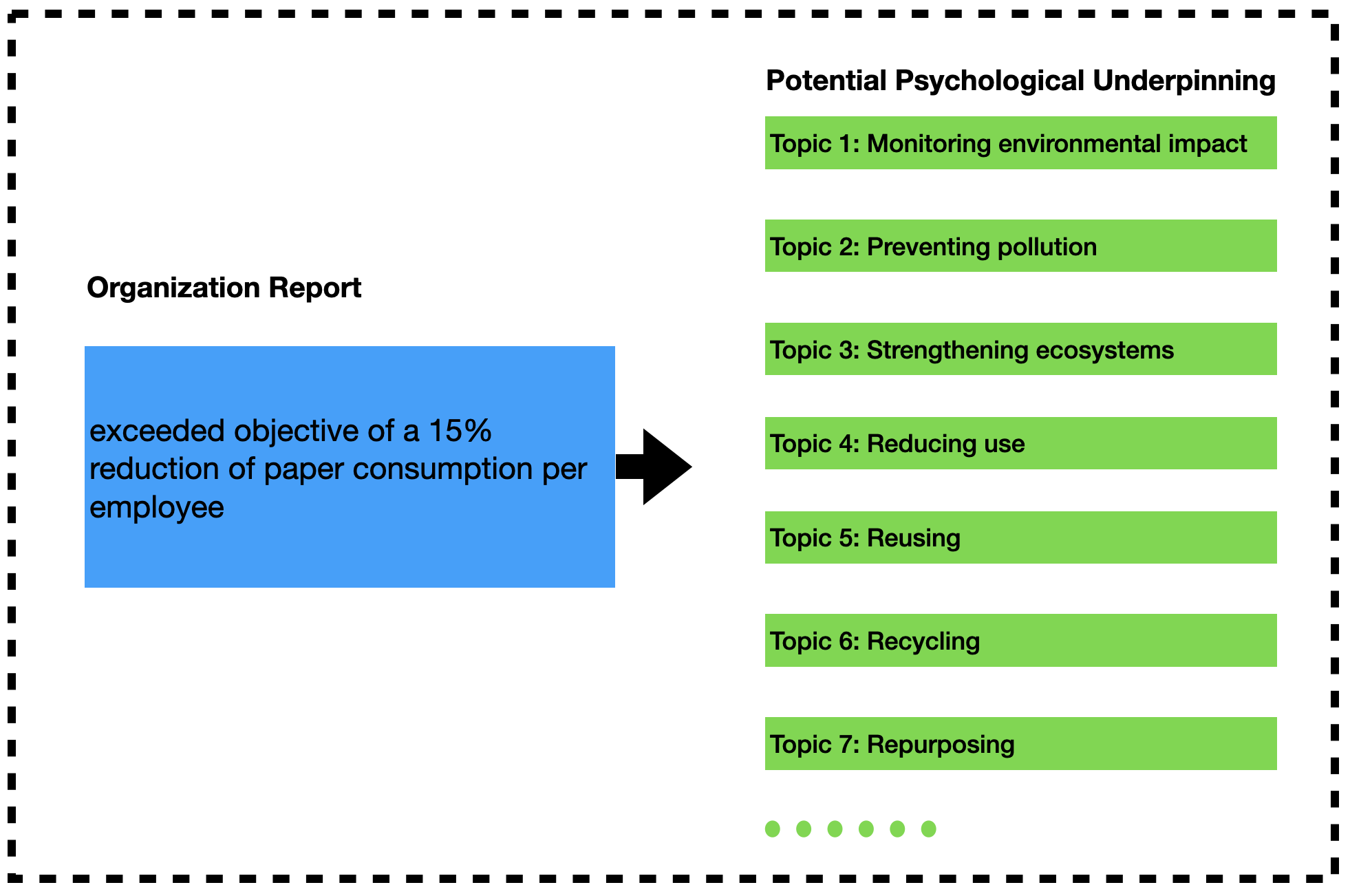}
\caption{Given an unstructured text, we transform it into one particular psychological underpinning through text analysis, specifically finetuning a large language model. In the provided example, the text is classified under the fourth topic,  reducing use.}
\label{multi}
\end{figure}

The objective of the research project is to understand the environmental endeavors of large organizations in China, Germany, France, the United Kingdom, and the Middle East and North Africa (MENA) regions by examining their self-reported initiatives detailed in sustainability reports~\citep{pm}. The taxonomy consists of the following 15 categories, each representing one major behavioral pathway through which an organization can pursue environmental sustainability: ``monitoring environmental impact'', ``preventing pollution'', ``strengthening ecosystems'', ``reducing use'', ``reusing'', ``recycling'', ``repurposing'', ``encouraging and supporting others'', ``educating and training for sustainability'', ``creating sustainable products and processes'', ``embracing innovation for sustainability'', ``changing how work is done'', ``choosing responsible alternatives'', ``instituting programs and policies'', ``others.''

\subsubsection*{Data preparation}
\noindent Data preparation is one of the three core steps in finetuning a language model. It gets the researchers' datasets into the right format through (1) splitting, (2) tokenization, and (3) formatting (Listing~\ref{lst:prep}).  In this exercise, the entire dataset has 9795 samples. Following~\cite{pm}, we keep 2449 samples (25\%) for testing. From the remaining 7346 samples, rather than use cross validation, we split them into a training set with 6121 samples (62.5\%) and a validation set with 1225 samples (12.5\%). Unlike the many methods used in~\cite{pm}, finetuning LLMs does not usually require manual feature engineering, such as the removal of stop words and lemmatization~\citep{qta,pm}. 

\begin{lstlisting}[language=Python, caption={Split the dataset into training, validation, and testing sets. Then encode them into tokens. No manual feature engineering is required.}, label={lst:prep}]

import pandas as pd
df = pd.read_csv("German_French_UK_China_MENA_040420.csv") # load data

def data_prep(df, seed):
  # 1. Split the dataset into train, dev, and test. 
  X_train, X_test, Y_train, Y_test = test_train_split(df, seed)

  X_train, X_dev, Y_train, Y_dev = train_test_split(X_train, Y_train, test_size=1225, random_state = seed, stratify = Y_train)

  X_train = list(X_train)
  X_dev = list(X_dev)
  X_test = list(X_test)

  Y_train = list(Y_train)
  Y_dev = list(Y_dev)
  Y_test = list(Y_test)

  # 2. Tokenize each dataset
  train_encodings = tokenizer(X_train, truncation=True, padding=True, max_length = 512) # set the max number of tokens in each input to 512
  dev_encodings = tokenizer(X_dev, truncation=True, padding=True, max_length = 512)
  test_encodings = tokenizer(X_test, truncation=True, padding=True, max_length = 512)

  # 3. Format the tokenized dataset
  train_dataset = PsyDataset(train_encodings, Y_train)
  dev_dataset = PsyDataset(dev_encodings, Y_dev)
  test_dataset = PsyDataset(test_encodings, Y_test)
  return train_dataset, dev_dataset, test_dataset

\end{lstlisting}

\subsubsection*{Finetuning a Large Language Model}

\noindent Once we get the datasets ready for use, we are ready for finetuning. The decisions we need to make here include: (1) which pretrained language model to use (large or small, RoBERTa or BERT, among others), (2) what hyperparameters we would like to use (learning rate, batch size, number of training epochs, etc).\footnote{Note that the number of epochs determines the number of times we want to train the model over the training set. The learning rate determines how much we adjust the model parameters during learning. A higher learning rate would mean we adjust the parameters more every time we learn from the training data. Batch size represents the number of samples we pack together in each model iteration.} In this exercise, we finetune a RoBERTa-large model~\citep{roberta} for text classification. RoBERTa-large has 24 layers of transformers and 340 million parameters in total. On top of its 24 layers of transformers, we add a classification layer for 15-topic classification. We use cross entropy as the loss function. We finetune the RoBERTa-large model for 10 epochs with a learning rate of 1e-5, a batch size of 64, and an input sequence length of 512 on an A100 GPU. We use the validation set's overall precision to select the best epoch and the optimal checkpoint. Interested readers could further adjust hyperparameters such as learning rate, number of training epochs and select the best hyperparameters based on the validation metrics. We then use the optimal checkpoint to make inferences on the test set with a batch size of 128 (see Listing~\ref{lst:multi}). We observe that it takes less than 20 minutes to finetune a RoBERTa-large over the training set for 10 epochs, which should fit into the time budget of most researchers.\footnote{The finetuning time would increase, for example, if we increase the number of training epochs or increase the size of the training set.}

\begin{lstlisting}[language=Python, caption={Specify the number of labels to be 15 for the multi-class classification and set the hyperparameters such as batch size and learning rate.}, label={lst:multi}]

# 1. Select the pretrained model to use
def model_init():
    return RobertaForSequenceClassification.from_pretrained("roberta-large", num_labels=15)

# 2. Set the hyperparameters to finetuning
training_args = TrainingArguments(
    output_dir="./llm-results",      # output directory
    num_train_epochs=10,             # total number of training epochs
    per_device_train_batch_size=64,  # batch size per device during training
    per_device_eval_batch_size=128,  # batch size for evaluation
    warmup_steps=0,                  # number of warmup steps for learning rate scheduler
    weight_decay=0.01,               # strength of weight decay
    logging_dir="./logs",            # directory for storing logs
    logging_steps=100,
    learning_rate = 1e-5,
    save_strategy= "epoch",
    evaluation_strategy="epoch",
    load_best_model_at_end= True,
    metric_for_best_model="precision",
    seed = 11,
)

# 3. Start finetuning with the selected model and the prepared datasets
trainer = Trainer(
  model_init=model_init,               
  args=training_args,                  # training arguments
  train_dataset=train_dataset,         # training dataset
  eval_dataset=dev_dataset,            # evaluation dataset
  )
trainer.train()

\end{lstlisting}

\subsubsection*{Model Evaluation}
\noindent After finetuning is completed, we need to evaluate how good the finetuned model is by performing evaluation on the test set. This step is the same for all models, whether they are trained from scratch as in~\cite{pm} or finetuned from a pretrained model as is done here. This step involves making predictions using the newly finetuned model and evaluating how accurate these predictions are, using metrics such as precision, recall and f1-score.\footnote{See https://scikit-learn.org/stable/modules/generated/sklearn.metrics.precision\_recall\_fscore\_support for the exact definitions of these metrics.} From a psychologist's perspective, this is where we transform the free-form texts into psychological constructs and evaluate how good these transformations are before putting the model to wider use.

\begin{lstlisting}[language=Python, caption={Evaluate the performance of the finetuned model on the test set.}, label={lst:multi}]
output = trainer.predict(test_dataset) # make predictions on the test set
metrics = classification_report(output.label_ids, np.argmax(output.predictions, axis=-1), output_dict = True) # calculate test metrics

precision = metrics[`weighted avg'].get(`precision')
recall = metrics[`weighted avg'].get(`recall')
f1 = metrics[`weighted avg'].get(`f1-score')
\end{lstlisting}

\subsubsection*{Classification Results}
\noindent In this subsection, we report the experiment results on classification. As our baselines, we report the original results by~\cite{pm}. In addition, recent research has shown that despite their impressive generative capabilities, generative large language models, such as ChatGPT, Gemini, and Llama, often lag behind fully finetuned BERT models for classification tasks~\citep{llm_transform_css,chatgpt_vs_bert,2024-navigating-prompt}. Therefore, we also report the zero-shot prompting results using ChatGPT-4o for comparison.\footnote{ChatGPT-4o api calls were made on July 11th, 2024. For the prompting details, please refer to our online appendix.}

In terms of performance, our finetuned RoBERTA-large model yields substantially superior results (Table~\ref{multi-class}). Across all metrics — precision, recall, and F1 — the finetuned RoBERTa model consistently outperforms earlier models, including Naïve Bayes, Support Vector Machine, and Max Entropy by~\cite{pm}, when trained with the same number of samples and evaluated using the same metrics. Specifically, in terms of F1 score, the finetuned RoBERTa (73.7\%) model outperforms the best existing model, Max Entropy, (65.4\%) by 13\%. In terms of precision, the finetuned LLM (73.9\%) outperforms Max entropy (65.7\%) by 12\%. Given the model's superior performance, researchers should feel comfortable to replace earlier models with the finetuned RoBERTa when we need to transform corporations' reports into those 15 psychological underpinnings.

\begin{table}[h]
\centering
\caption{A finetuned RoBERTa-large model outperforms Naïve Bayes, Support Vector Machine, Max Entropy, and Zero-Shot Prompting with ChatGPT-4o in the 15-topic classification task by a large margin. Results on Naïve Bayes, Support Vector Machine, and Max Entropy are from~\cite{pm}. Means of five random runs for the finetuned LLM and for zero-shot prompting are reported. Better results are in bold and indicating a higher capability of transforming texts into psychological constructs.}
\renewcommand{\arraystretch}{1.1}
\label{multi-class}
\begin{tabular}{@{\extracolsep{25pt}}lccc@{}}
\hline\hline
Model          & Precision & Recall & F1   \\
\hline
Naïve bayes   & 0.624     & 0.589  & 0.567   \\
Support vector & 0.652     & 0.651  & 0.650   \\
Max entropy    & 0.657     & 0.656  & 0.654   \\
Zero-shot prompting & 0.492 & 0.470 & 0.455 \\ 
\textbf{Finetuned LLM}  & \textbf{0.739}     & \textbf{0.738}  & \textbf{0.737}  \\
\hline
\end{tabular}
\end{table}

\subsection{Regression: Comparing Language Model Fine-tuning with the \textit{Text} Package in R}

\begin{figure}[h]
\centering
\includegraphics[width=425px]{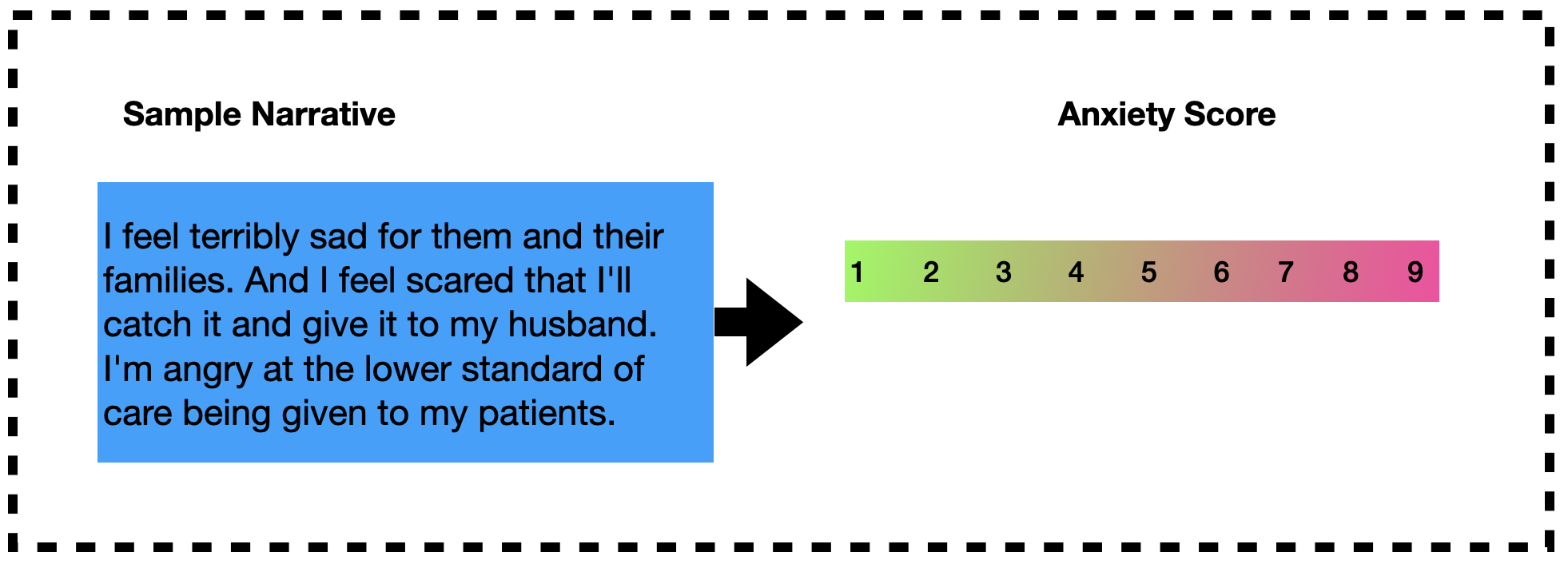}
\caption{Given an unstructured text, we transform it into one particular anxiety level through text analysis, specifically finetuning a large language model. In the provided example, the text snippet has an anxiety score of 8.}
\label{regression-sample}
\end{figure}

\noindent In the next case study, we transform written narratives into anxiety scores between 1 and 9 (Figure~\ref{regression-sample}). Our goal is two-fold: to show how regression with fine-tuned language models can be done with relative ease and to show that finetuning a large language model can be more effective than merely using language models for embedding as is done in the \textit{Text} package~\citep{text_package}. Data for this regression experiment was sourced from the Real World Worry Dataset (RWWD) by~\cite{kleinberg2020measuring}. The RWWD contains text data and corresponding emotion scores from nine categories—anger, anxiety, desire, disgust, fear, happiness, relaxation, sadness, and worry—from 2,500 individuals during the initial phase of the COVID-19 pandemic. Participants self-reported each emotion, including the dominant one, rating them on a 9-point scale. Unlike typical text analysis datasets reliant on third-party annotations, the RWWD utilizes a direct survey method to gather these self-reported emotions and written narratives. This approach is especially advantageous for psychological research as it offers direct insights into participants' immediate thoughts and feelings, thereby minimizing the biases usually associated with external interpretations. Moreover, this methodology proves invaluable for text mining, providing rich, authentic datasets essential for detailed analysis of linguistic and emotional expression. Notably, anxiety emerged as the predominant emotion, thus becoming a focal point for our text-mining endeavors. Consequently, we focus on analyzing anxiety levels alongside lengthy text entries (minimum 500 characters) for the regression tasks.

\subsubsection*{Data preparation}

\noindent In Table~\ref{anxiety}, we report the distribution of the Anxiety dataset. In total, we have 2500 labeled samples. We sample 500 to use as the test set and keep the remaining 2000 as the training set. From the training set, we further reserve 10\% as the validation set (Listing~\ref{lst:prep_b}). There are nine different levels of anxiety in total, from 1 to 9, with Levels 7, 8, and 9 being the most common.

\begin{table}[h]
\centering
\renewcommand{\arraystretch}{1.1}
\begin{tabular}{lclllllllll}\hline\hline
\multirow{2}{*}{Dataset} & \multirow{2}{*}{Number of samples} & \multicolumn{9}{c}{Level of Anxiety}               \\\cline{3-11}
                         &                                    & 1  & 2   & 3   & 4   & 5   & 6   & 7   & 8   & 9   \\\hline
Training set             & 2000                               & 78 & 116 & 102 & 115 & 129 & 246 & 384 & 456 & 374 \\
Test set              & 500                                & 22 & 22  & 24  & 22  & 31  & 58  & 92  & 134 & 95  \\\hline\hline
\end{tabular}
\caption{Summary statistics of the \textit{Anxiety} dataset.}\label{anxiety}
\end{table}

\begin{lstlisting}[language=Python, caption={Similar to the classification exercise, we split the dataset into training, validation, and test sets. Then encode them into tokens. No manual feature engineering is required.}, label={lst:prep_b}]

column = "anxiety" # the y column
mlength = 512 # max number of tokens
train = pd.read_csv("/content/train_dev.csv") # load the datasets
test = pd.read_csv("/content/test.csv")
train[column] = train[column].astype("float32") # conver the anxity column into float32
test[column] = test[column].astype("float32")

def train_dev_split(df, seed): # only split the train and dev sets
  X_train, X_dev, Y_train, Y_dev, indices_train, indices_test = train_test_split(df["text_long"], df[column], df.index, test_size=len(df)//10, random_state = seed)
  return [X_train, X_dev, Y_train, Y_dev] # effectively save 10% as validation

def data_prep(df, seed = 11): # set the default seed to 11
  X_train, X_dev, Y_train, Y_dev = train_dev_split(df, seed)

  X_train = list(X_train)
  X_dev = list(X_dev)
  X_test = list(test["text_long"])

  Y_train = list(Y_train)
  Y_dev = list(Y_dev)
  Y_test = list(test[column])

  train_encodings = tokenizer(X_train, truncation=True, padding=True, max_length=mlength)
  dev_encodings = tokenizer(X_dev, truncation=True, padding=True, max_length = mlength)
  test_encodings = tokenizer(X_test, truncation=True, padding=True, max_length= mlength)

  train_dataset = PsyDataset(train_encodings, Y_train)
  dev_dataset = PsyDataset(dev_encodings, Y_dev)
  test_dataset = PsyDataset(test_encodings, Y_test)
  return train_dataset, dev_dataset, test_dataset

\end{lstlisting}

\subsubsection*{Finetuning a Large Language Model}
\noindent Next, we finetune a BERT-base-uncased model~\citep{bert} for text regression.\footnote{In addition to finetuning the BERT-base-uncased model as described in the code snippet, we also finetune a RoBERTa-base model. In the results subsection, we report results from both models.} BERT-base has 12 layers of transformers and 110 million parameters in total. On top of the 12 layers of transformers, we add a regression layer and use mean squared error as the loss function. We finetune the pretrained BERT-base model for 10 epochs with a learning rate of 2e-5, a batch size of 32 for training, and a batch size of 64 for inference. Some of the key differences that distinguish regression from classification include (1) the number of labels is set to 1 and (2) the loss functions that we use for training the model and for selecting the optimal checkpoints.

\begin{lstlisting}[language=Python, caption={Specify the number of labels to be 1 for the regression task and set the hyperparameters such as batch size and learning rate.}, label={lst:train_b}]
training_args = TrainingArguments(
  output_dir="./llm-results",        # output directory
  num_train_epochs=10,               # total number of training epochs
  per_device_train_batch_size=32,    # batch size per device during training
  per_device_eval_batch_size=64,     # batch size for evaluation
  warmup_steps=0,                    # number of warmup steps
  weight_decay=0.01,                 # strength of weight decay
  logging_dir="./logs",              # directory for storing logs
  logging_steps=100,
  learning_rate = 2e-5,
  save_strategy= "epoch",
  evaluation_strategy="epoch",
  load_best_model_at_end= True,
  metric_for_best_model="mse",
  greater_is_better=False,
  seed = 11,
)

def model_init():
    return BertForSequenceClassification.from_pretrained("bert-base-uncased", num_labels=1)

# 3. Start finetuning with the selected model and the prepared datasets
train_dataset, dev_dataset, test_dataset = data_prep(train[:size])
trainer = Trainer(
  model_init=model_init,               # the instantiated Transformers model to be trained
  args=training_args,                  # training arguments, defined above
  train_dataset=train_dataset,         # training dataset
  eval_dataset=dev_dataset,            # evaluation dataset
  compute_metrics=compute_metrics,     # compute_metrics
  )
trainer.train()
\end{lstlisting}

\subsubsection*{Model Evaluation}
\noindent Once the finetuning is completed, we evaluate the regression model using metrics such as mean squared error and Pearson's R. Most of these metrics are readily available from existing libraries and we just need to load them to calculate the metrics that we need (Listing~\ref{regress_3}). Intuitively, for model evaluation researchers need the ground truth as well as the predictions from the newly finetuned model. With these two lists of numbers, we then proceed to use existing libraries to calculate relevant metrics, such as mean squared error.

\begin{lstlisting}[language=Python, caption={Evaluate the performance of the finetuned model on the test set.}, label={regress_3}]
  trainer.train()
  output = trainer.predict(test_dataset)
  mse_metric = load("mse")
  pearsonr_metric = load("pearsonr")
  m = mse_metric.compute(predictions=output.predictions, references=output.label_ids, squared = False)
  p = pearsonr_metric.compute(predictions=output.predictions, references=output.label_ids)
\end{lstlisting}

\subsubsection*{Regression Results}
\noindent We report our experiment results for regression using the \textit{Anxiety} dataset. As our baseline and comparison group, we use transformers as the embedding models and train regression models using text embeddings. This approach represents an alternative usage of transformer models~\citep{language_marker,depression_prediction,promotional_language}.\footnote{Yet another usage of transformers is to use embeddings from the transformers for clustering and similarity measurement~\citep{linguistic_agency}.} Specifically, we use the \textit{Text} package released by~\cite{text_package}. In the \textit{Text} package, the core function is \textbf{textEmbed}, which converts natural language into word embeddings using transformer models such as BERT and RoBERTa. These embeddings then serve as the input for further functions like \textbf{textTrain} and \textbf{textSimilarity}, enabling various downstream tasks including regression.

For training models, we progressively sample 500, 1000 and 2000 data points. The \textit{Text} package~\citep{text_package} uses cross-validation and trains a ridge regression model~\citep{elements} by default. By contrast, for fine-tuning language models, we sample 10\% of the training samples for validation to select the optimal checkpoint. We mostly use the default values for both the \textit{Text} package and fine-tuning.


In Table~\ref{anxiety_rating_results}, we report the results on regression. We observe that for the regression task on the \textit{Anxiety} dataset, fine-tuning language models consistently outperforms the \textit{Text} package across all metrics: Pearson's \textit{r}, RMSE, and running time. With 2000 training samples, the fine-tuned RoBERTa model has a Pearson's \textit{r} at 0.602 and is 28\% higher than the Pearson's \textit{r} by the~\textit{Text} package where the same RoBERTa model is used but restricted to generating embeddings. We further note that fine-tuning a transformer model, BERT or RoBERTa, with 500 samples can be more effective and faster than training the same transformer-based model in~\textit{Text} package with 2000 samples.\footnote{Note that another observation is that the finetuned model tends to perform better with more training data (500, 1000, 2000). While it is generally true that the more data the better, the performance gain from an increase in data size diminishes as we add more and more data.}

\begin{table}[h]
\setlength{\tabcolsep}{7.5pt}
\renewcommand{\arraystretch}{1.2}
\begin{tabular}{llllllll}
\hline\hline
       &   & \multicolumn{6}{c}{Anxiety Rating}                                                                      \\\cline{3-8} 
        &  & \multicolumn{2}{c}{500 samples} & \multicolumn{2}{c}{1000 samples} & \multicolumn{2}{c}{2000 samples} \\ \hline
    Model  &  Method & \textit{Text} & fine-tune & \textit{Text} & fine-tune & \textit{Text} & fine-tune \\ \hline
  \multirow{3}{*}{BERT}     &  Pearson's \textit{r}  & 0.446 & \textbf{0.550} & 0.464 & \textbf{0.558}  &0.483 & \textbf{0.589}\\ 
      &  RMSE &  2.018 & \textbf{1.958} & 1.999 & \textbf{1.945} & 1.980 & \textbf{1.846} \\
      &  Time  & 31m & \textbf{222s} & 60m & \textbf{7m} & 120m & \textbf{14m} \\ \hline
   \multirow{3}{*}{RoBERTa}    &  Pearson's \textit{r}  & 0.426 & \textbf{0.507} & 0.454 & \textbf{0.570}  &0.470 & \textbf{0.602}\\ 
      &  RMSE &  2.046 & \textbf{1.982} & 2.011 & \textbf{1.924} & 1.996 & \textbf{1.812} \\
      &  Time  & 14m & \textbf{199s} & 27m & \textbf{8m} & 53m & \textbf{14m} \\ \hline

\end{tabular}
\caption{Performance comparison between the \textit{Text} package and fine-tuning for regression. Better results (higher correlation, lower RMSE and less time) in bold. fine-tuning consistently outperforms the \textit{Text} package in terms of Pearson's \textit{r}, RMSE, and running time.}
\label{anxiety_rating_results}
\end{table}

We further note that for the fine-tuned RoBERTa model performs better than the fine-tuned BERT model when there are 1000 or more training samples. For example, with 2000 training samples, 10\% being reserved for validation, the fine-tuned BERT has a linear correlation score of 0.589 as compared with the fine-tuned RoBERTa's 0.602. This suggests that researchers should consider experimenting with more than one transformer model and pick the one that performs best given a particular task.\footnote{For a list of existing models, readers could start with https://huggingface.co/models.}

In terms of running time, fine-tuning BERT and fine-tuning RoBERTa takes roughly the same amount of time on an A100 GPU.\footnote{Note that it is technically not fair to compare running time on a CPU with the running time on a GPU. The purpose of our comparison is more in practical terms: how much time researchers would need if they train a model using the \textit{Text} package versus how much time the researchers need if they fine-tune the same language model instead.} They are substantially faster than running the \textit{Text} package on a CPU. In the case with 2000 training samples and the BERT model, running the \textit{Text} package on a CPU can take almost 9 times as much time as fine-tuning the same transformer model on a GPU does.

\section{Conclusion}
Text analysis plays an important role in psychological research by extracting meaning from free-form text. Such analysis can be done through human coding, training small models from scratch with lots of data, and more recently, finetuning large models with a relatively small amount of data. This latest approach, known as the pretrain-finetune paradigm, has revolutionized the field of natural language processing. It makes it possible for researchers to achieve competitive results with relatively few labeled samples, all while being remarkably easy to use. In this tutorial, we have provided an intuitive and thorough walk-through of the key concepts therein. We also introduced some of the most common use cases that the pretrain-finetune paradigm supports. To help facilitate the wider adoption of this new paradigm, we have included easy-to-follow examples and made the related datasets and scripts publicly available. Practitioners and scholars in psychology and social science broadly should find our tutorial useful.

\newpage
\bibliographystyle{apacite}
\bibliography{ir}
\end{document}